\pdfoutput=1 
\documentclass[11pt, a4paper]{article}
\usepackage[utf8x]{inputenc}
\usepackage[T1]{fontenc}
\usepackage{graphicx}
\usepackage{float}
\usepackage{soul}
\usepackage{textcomp}
\usepackage[nointegrals]{wasysym}
\usepackage{latexsym}
\usepackage{amsmath}
\usepackage{amssymb}
\usepackage{makecell}
\usepackage[normalem]{ulem}
\usepackage{xcolor}
\usepackage[hyperfootnotes=false]{hyperref}
\usepackage[hyperref]{acl2019}
\aclfinalcopy

\usepackage{amsmath} \usepackage{ragged2e} \usepackage{soul} \usepackage{setspace} \usepackage{multirow} \usepackage{times} \usepackage{latexsym}
\author{William Léchelle, Fabrizio Gotti, Philippe Langlais \\
  RALI, University of Montreal \\
  \texttt{\{lechellw, gottif, felipe\} @iro.umontreal.ca}
}

\date{}
\title{WiRe57 : A Fine-Grained Benchmark for Open Information Extraction}
\hypersetup{
pdftitle={WiRe57 : A Fine-Grained Benchmark for Open Information Extraction},
 pdfauthor={William Léchelle, Fabrizio Gotti, Philippe Langlais},
 pdfkeywords={open information extraction ; evaluation},
 pdfsubject={},
 pdflang={English}}
\begin{document}

\maketitle
\begin{abstract}
   We build a reference for the task of Open Information Extraction, on five
   documents. We tentatively resolve a number of issues that arise, including
   coreference and granularity, and we take steps toward addressing inference, a significant problem. We seek to better pinpoint the requirements for
   the task. We produce our annotation guidelines specifying what is correct to
   extract and what is not. In turn, we use this reference to score existing
   Open IE systems. We address the non-trivial problem of evaluating the
    extractions produced by systems against the reference
   tuples, and share our evaluation script. Among seven compared extractors, we
   find the MinIE system to perform best.
\end{abstract}

\section{Introduction}
\label{sec:orgb2105d7}
Open Information Extraction systems, starting with TextRunner
\cite{Yates:2007:TOI:1614164.1614177}, seek to extract all relational tuples
expressed in text, without being bound to an anticipated list of predicates.
Such systems have been used recently for relation extraction
\cite{DBLP:conf/tac/SoderlandGBEW13}, question-answering
\cite{Fader:2014:OQA:2623330.2623677}, and for building domain-targeted
knowledge bases \cite{DBLP:journals/tacl/MishraTC17}, among others.

Subsequent extractors (ReVerb, Ollie, ClausIE, OpenIE 4, etc.) have sought
to improve yield and precision, i.e. the number of facts extracted from a
given corpus, and the proportion of those facts that is deemed correct.

Nonetheless, the task definition is underspecified, and, to the best of our
knowledge, there is no gold standard. Most evaluations require somewhat
subjective and inconsistent judgment calls to be made about extracted tuples
being acceptable or not.  The most recent automatic benchmark of
\citet{Stanovsky2016EMNLP} has some shortcomings that we propose to tackle
here, regarding the theory underlining the task definition as well as the
evaluation procedure.

We manually performed the task of Open Information Extraction on
5~short documents, elaborating tentative guidelines for the task, and resulting
in a ground truth reference of 347 tuples. We evaluate against our benchmark
the available OIE engines up to MinIE, with a fine-grained token-level
evaluation. We distribute our resource and annotation guidelines, along with
the evaluation script.\footnote{\url{https://github.com/rali-udem/WiRe57}}


\section{Related Work}
\label{sec:org385e7da}
For their evaluation, typically, developers of Open IE systems pool the
output of various systems on a given corpus. They label a sample of produced
tuples as correct or incorrect, with the general guideline that an
extraction is correct if it is implied by the sentence. Thus,
\newcite{ollie-emnlp12} write: \emph{``Two annotators tagged the extractions as
correct} \emph{if the sentence asserted or implied that the relation was true.''}
\newcite{DelCorro:2013:CCO:2488388.2488420} propose: \emph{``We also asked
labelers to be liberal with respect to coreference or entity} \emph{resolution;
e.g., a proposition such as (`he' ; `has' ; `office'), or} \emph{any unlemmatized
version thereof, is treated as correct.}'' \newcite{DBLP:conf/acl/SahaPM17}:
\emph{``We sample a random testset of 2,000 sentences} \emph{[\ldots{}]} \emph{Two annotators
with NLP experience annotate each extraction for correctness.''}
\citet{D17-1278}: \emph{``A triple is labeled as correct if it is entailed by its
corresponding clause.''} Then, precision and yield are used as performance
metrics. Without a reference, recall is naturally impossible to measure.

We define a reference \emph{a priori}. This allows for automatic scoring of
systems' outputs, which greatly diminishes subjectivity from the process of
labelling facts ``for correctness''. Above all, it is meant to help
researchers agree on what the task precisely entails. Therefore, it
allows to measure a true recall (albeit on a small corpus).

The complexity of our guidelines is indicative of all that is swept under
the carpet when ``annotating for correctness''. As a matter of fact, when
closely examining other references for OIE, many extracted tuples eventually
labelled as ``good'' have more or less important issues. Some really dubious
cases are hard to gauge and their labelling is ultimately subjective. To
showcase the devilishly difficult judgment calls that this implies, compare
the following two extractions. ``\textit{`The opportunity is
   significant and I hope we can take the opportunity to move forward,' he said
   referring to his coming trip to Britain.}'' yields (\emph{his ; has ; coming
trip}), and ``\textit{[...], the companies included CNN, but not its parent, AOL Time Warner}'' yields (\emph{its~; has ; parent}). Are the
extractions implied by the sentence~? In
\cite{DelCorro:2013:CCO:2488388.2488420}, the annotator approved the latter,
and rejected the former. The extraction (\emph{he ; said} ; \emph{The opportunity is
significant referring to his coming trip to Britain}) was also deemed
correct, despite the composed second argument.

Some other tasks for which OIE output is used, such as Open QA
\cite{Fader:2014:OQA:2623330.2623677}, TAC-KBP
\cite{DBLP:conf/tac/SoderlandGBEW13}, or textual similarity and reading
comprehension as in \cite{stanovsky-dagan:2015:ACL-IJCNLP} --- could in
principle be used to compare extractors' performance, but only give a very
coarse-grained signal, mostly unaffected by the tuning of systems.

A promising method is that explored by \citet{DBLP:journals/tacl/MishraTC17}
for the Aristo KB.\footnote{\url{http://data.allenai.org/tuple-kb/}} Aristo is a
science-focused KB extracted from a high-quality 7M-sentence corpus. The
authors preprocessed a smaller, similarly science-related, independent
corpus of 1.2M sentences, into a "Reference KB" of 4147 facts, validated by
Turkers. Assuming that these 4147 facts are representative of the science
domain as a whole, they measured comprehensiveness (recall) over this domain
by measuring coverage on the Reference KB.

\subsection{ORE benchmark}
\label{sec:orgea721fe}

\citet{mesquita-schmidek-barbosa:2013:EMNLP} compare more or less deep
`parsers', including the OIE systems Ollie and ReVerb, on the germane task
of Open Relation Extraction (ORE), between named entities. They build a benchmark
of 662 binary relations over 1100 sentences from 3 sources (the Web, the
New York Times and the Penn Treebank). They label an additional 222 NYT
sentences with as many $n$-ary relations, and 12,000 with
automatic annotations. 

Besides the named entity arguments, their annotations consist of one
mandatory trigger word (indicating the relation), surrounded by a window of
allowed tokens. To compare OIE with ORE systems, they have to replace the
target entities by salient arguments (\emph{Asia} and \emph{Europe}) which are easy
to recognize. They discuss some of the challenges that arise from divergent
annotation styles and evaluation methods.

While the tasks are similar, restraining arguments to be named entities
limits IE to capturing only the most salient relations expressed in the
text. Allowing for any NP to be an argument, we extract 6 facts per
sentence on average in the benchmark presented here, compared to 0.6 in
the ORE dataset. We also annotate some relations that do not have a trigger
in the sentence (such as (\emph{Paris~; [is in] ; France}) from \emph{``Chilly Gonzales
lived in Paris, France''}).
\subsection{QA-SRL OIE benchmark}
\label{sec:org4aff63d}
\newcite{Stanovsky2016EMNLP} build a large benchmark for OIE, by automatically
processing the QA-SRL dataset \cite{D15-1076}. Precisely, for each predicate
annotated in QA-SRL, they generate one tuple expressing each element of the
Cartesian product of answers to the questions about this
predicate.  

For instance, QA-SRL lists five questions asked about the sentence \emph{``Investors
are appealing to the SEC not to limit their access to} \emph{information about stock
purchases and sales by corporate insiders}''~: \emph{``who are \uline{appealing} to
something ?}'', \emph{``who are someone \uline{appealing} to ?}'', \emph{``what are someone
\uline{appealing} ?}'', \emph{``what might not \uline{limit} something ?}'' and \emph{``what might
not someone \uline{limit} ?}'', with one answer per question. This generates the
reference tuples (\emph{Investors ; appealing ; not to limit their access to
information} \emph{about stock purchases and sales by corporate insiders ; to the
SEC}) and (\emph{the SEC ; might not limit} ; \emph{their access to information about
stock purchases and sales by corporate insiders}).

Their dataset is comprised of 10,359 tuples over 3200 sentences (from the Wall
Street Journal and Wikipedia), and is available for
download.\footnote{\url{http://u.cs.biu.ac.il/\~nlp/resources/downloads/}} 

While this work makes a big step in the right direction, there are a few
important issues with this benchmark. 

First, a major strength of the dataset is its intended and partly achieved
completeness, but we do not find it to be a suitably comprehensive reference
against which to measure systems' recall. This might be because the QA-SRL
dataset doesn't lend itself well to exhaustiveness in the realm of Open IE,
partly because it is restricted to explicit predicates.  For instance, the
sentence \emph{``However, Paul Johanson, Monsanto's director of plant sciences, said
the company's chemical spray} \emph{overcomes these problems and is `gentle on the
female organ'.}'' contains two predicates, generating the extractions (\emph{Paul
Johanson ; said} ; \emph{the company's chemical spray overcomes these problems and
is ``gentle on the female organ.''}) and (\emph{the company's chemical spray ;
overcomes ; these problems}). Yet, that omits the (in our view useful)
extractions (\emph{the company's chemical spray ; is ; ``gentle on the female
organ''}), and (\emph{Paul Johanson ; is ; Monsanto's director of plant sciences}).

Another issue is that some words not found in the original sentence were quietly added
by the SRL-to-QA process, retained in the QA-to-OIE transformation, and
become part of the reference. In the example above, it is unclear how the
second predicate \emph{``might not limit}'' is extracted from the sentence. At the very least, the fact that these words are foreign to the original sentence should be made explicit. Further,
although in this particular case adding the modal is a good way of expressing
the information, its repeated use by QA-SRL to produce questions waters down
the expressed facts in the end. For instance, the uninformative triple (\emph{a
manufacturer ; might get ; something}) is generated from the sentence \emph{``\ldots{}and
if a manufacturer is clearly trying to get something out of it \ldots{}}'', with the
same added \emph{``might}''.

\begin{figure}[ht!]
  \centering
  \makebox[100pt]{
    \includegraphics[scale=.43]{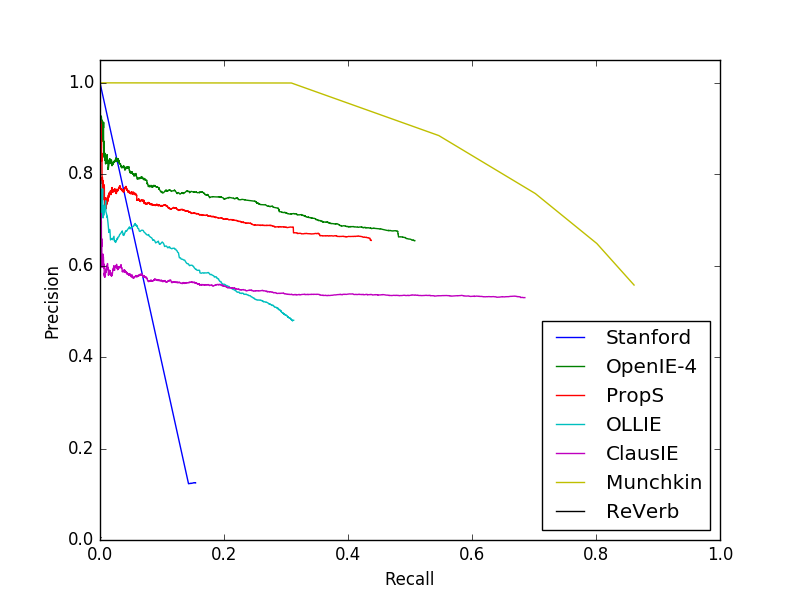}
  }
  \caption{Performance metrics must take span precision into account. The
    25-line long Munchkin script returns variations of the full sentence (with
    decreasing confidence) and is not penalized by the evaluation script of the
    latest benchmark \cite{Stanovsky2016EMNLP}. Its superior performance is
    artificially inflated.}
  \label{fig-munchkin}
\end{figure}

Last, the scoring procedure is not robust. Using the code made available by the
authors\footnote{\url{https://github.com/gabrielStanovsky/oie-benchmark} --- the scoring
function was updated since its description in the article. We
believe the published function suffers from similar issues.}, we were
able to get top results with a dummy extractor.

This is because the scorer doesn't penalize extractions for being too long, nor for
misplacing parts of the relation in the object slot or vice versa. Therefore,
if \(w_0 w_1 ... w_n\) is an input sentence, a trivial system that "extracts"
\((w_0 ; w_1 ; w_2...w_n)\), \((w_0 ; w_1 w_2 ; w_{3}...w_n)\), etc., will be given
an unfairly great score. We implemented that program (dubbed Munchkin) which
predictably performed well above other genuine extraction systems, as pictured
in Figure \ref{fig-munchkin}.

\subsection{RelVis benchmarking toolkit}
\label{sec:org80db224}
\citet{W17-5402} evaluate four systems (ClausIE, OpenIE 4, Stanford Open IE and
PredPatt) against the two datasets mentioned above.\footnote{Their code is announced but not
available as of this writing --- \url{https://github.com/schmaR/relvis}} They
use two methods to match predicted and reference tuples : ``containment'' and
``relaxed containment''.  These methods mean that the predicted tuple must
include the reference tuple, and that inclusion must happen for each argument,
in the non-relaxed case. In the relaxed case, the boundaries between parts of a
tuple are ignored. Like that of \citet{Stanovsky2016EMNLP}, this scoring
procedure doesn't penalize systems for returning overlong spans.
\subsection{Scoring}
\label{sec:orgdb42d39}
To compare facts with a reference, most authors require matching tuples to
have the same number of arguments and to share the grammatical head words
of each part, e.g. \citet{DBLP:conf/conll/AngeliM13} and the article
of \citet{Stanovsky2016EMNLP}. In their updated GitHub repository, \citet{Stanovsky2016EMNLP} instead use lexical match : more than half
of the words of a predicted tuple must match the reference for it to be correct. 

In contrast with these works and \cite{W17-5402}, our scorer penalizes
verbosity by measuring precision at the token level. We penalize the
omission of parts of a reference tuple by gradually diminishing recall (at
the token level), instead of a sharp all-or-nothing criterion.

\citet{mesquita-schmidek-barbosa:2013:EMNLP} annotate relations as one
mandatory target plus some optional complementary words, and treat
arguments (named entities) in an ad-hoc fashion for OIE systems.

\section{WiRe57}
\label{sec:orgdb6a137}
Open IE bears some similarity to the task of Semantic Role Labelling, as
explored in \cite{Christensen:2011:AOI:1999676.1999697,mesquita-schmidek-barbosa:2013:EMNLP}, and as demonstrated by SRLIE, a
component of OpenIE~4. 

In effect extracted tuples are akin to simplified PropBank\footnote{\url{propbank.github.io} -- \citep{Kingsbury02fromtreebank}} or FrameNet\footnote{\url{framenet.icsi.berkeley.edu} -- \citep{framenet}} frames, and our annotations were inspired by those
projects. Still, with a focus on extracting new relations at scale, optional
arguments such as Propbank's modifiers (ArgM) are \emph{discouraged} in
OIE. Another major difference is the vocabulary of predicates being open to
any relational phrase, rather than belonging to a closed curated list such
as VerbNet. Within reason, OIE seeks to extract rich and precise relations
phrases. 

\begin{table}[htb]
\centering
\begin{tabular}{lrr}
Phenomenon & N & \%\\
\hline
All tuples & 343 & 100\\
Anaphora & 196 & 57\\
Contains inferred words & 186 & 54\\
Hallucinated parts & 135 & 39\\
Binary relations & 254 & 74\\
$n$-ary, \(n\) = 3 & 72 & 21\\
$n$-ary, \(n\) = 4 & 16 & 5\\
$n$-ary, \(n\) = 5 & 1 & 0.3\\
\hline
Inferred words & 347/2597 & 13.4\\
\end{tabular}
\caption{\label{tab:org109a930}
Frequencies of various phenomena in WiRe57.}
\end{table}

\subsection{Annotation process}
\label{sec:org85568e7}
A small corpus of 57 sentences taken from the beginning of 5~documents in
English was used as the source text from which to extract tuples. Three
documents are Wikipedia articles (Chilly Gonzales, the EM algorithm, and Tokyo)
and two are newswire articles (taken from Reuters, hence the Wi-Re name).

Two annotators (authors of this paper) first independently extracted tuples
from the documents, based on a first version of the annotation guidelines
which quickly proved insufficient to reach any significant agreement. The
two sets of annotations were then merged, and the guidelines rectified
along the way in order to resolve the issues that arose. After merging, a quick test on a few additional sentences from a different document showed a much improved agreement, more than half of extractions matching exactly and the remaining missing a few details. The guidelines are detailed in the next sections.

\subsection{Annotation principles}
\label{sec:org1c2d80b}
In keeping with past literature, our guiding principles for the annotation
were as follows.

The first, obvious purpose of extracted information is to be
\textbf{informative}. \citet{Fader:2011:IRO:2145432.2145596} mention how
extracting (\emph{Faust ; made ; a deal}) instead of the correct (\emph{Faust ; made
a deal with ; the devil}) would be pointless. Further, anaphoric mentions
being so ubiquitous and being void of meaning outside the context of their
original sentence, we resolve anaphora in our extractions.

Moreover and following \cite{Stanovsky2016EMNLP}, extracted tuples should
each be \textbf{minimal}, in the sense that they should convey the smallest
standalone piece of information, though that piece must be completely
expressed. Thus, some facts must be extracted as
$n$-ary relations.\footnote{Some systems --- namely CSD-IE
\cite{bast2013open} and NestIE \cite{D16-1006} --- explore nesting
extractions, but we didn't adopt this strategy.} The MinIE system in particular addresses this issue and ``minimizes its
extractions by identifying and removing parts that are considered overly
specific''.

The annotation shall be \textbf{exhaustive}, in the sense of capturing as much of
the information expressed in the text as possible. This is to measure
absolute recall for a system, a notoriously difficult evaluation metric for
Open~IE.

This in turn raises the issue of \textbf{inference}: some information is merely
suggested by the text, rather than explicitly expressed, and should not be
annotated. Light inference, in the form of reformulation, is helpful to
make use of the information extracted, but full-fledged inference should be
processed by a dedicated program, and is not part of the Open IE task. Because the concept of ``light inference'' is subjective, we propose in the guidelines a few examples and counterexamples that delineate the limits between the two classes. 

Other authors mention this issue. From \cite{Wu:2010:OIE:1858681.1858694} :
``The extractor should produce one triple for every relation stated
explicitly in the text, but is not required to infer implicit facts.''
\citet{Stanovsky2016EMNLP} say: ``an Open IE extractor should produce the tuple
(\emph{John; managed to open; the door}) but is not required to produce the
extraction (\emph{John; opened; the door})''. In our resource we do also
annotate (\emph{John; [opened]; the door}), marking the reworded relation as
inferred (which in turn makes it optional to find when scoring).

\begin{figure}[htb]
\centering
  {\includegraphics[scale=.4]{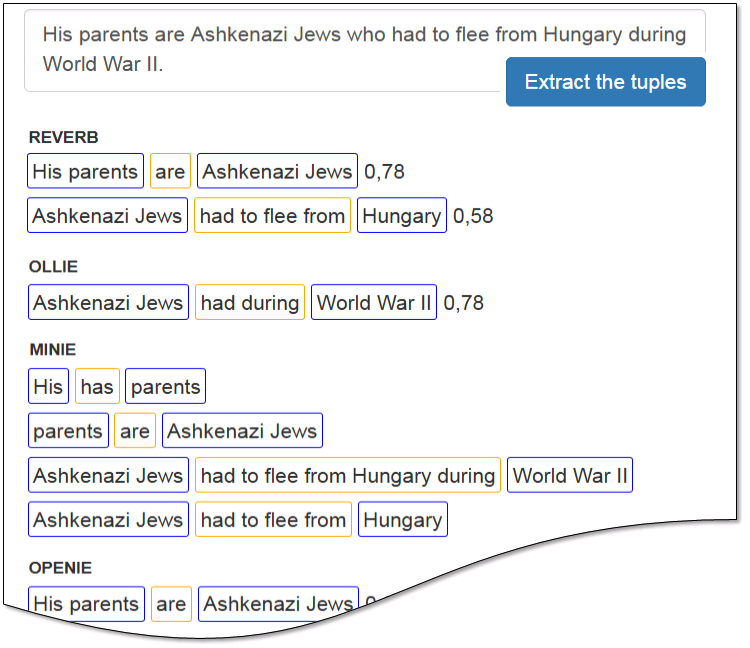}}
  \caption{Example output of evaluated OIE systems, on sentence CH 7. This cropped 
    screenshot is of a in-house web application that allows us to submit any
    sentence for tuple extraction and to visualize the results.}
  \label{fig-buddleia}
\end{figure}

\begin{figure}[htb!]
\small
\textbf{Sentence CH 7 --} ``\textit{His parents are Ashkenazi Jews who had to flee from
Hungary during World War II.}''

\smallskip\textbf{Annotations}
\\-- (His/(Chilly Gonzales's) parents ; are ; Ashkenazi Jews)
\\-- (His/(Chilly Gonzales's) parents ; are ; Jews)
\\-- ({His/(Chilly Gonzales's) parents ; had to flee from~;  \\\-\hspace{100pt} Hungary~; during World War II})
\\-- ({His/(Chilly Gonzales's) parents ; [fled] from ; Hungary~; \\\-\hspace{142pt} during World War II})
\\-- ([Chilly Gonzales] ; [has] ; parents)
\\\smallskip\hrulefill

\textbf{Sentence EM 5 --} ``\textit{They pointed out that the method had been `proposed
many times in special circumstances' by earlier authors.}''
\smallskip\\\textbf{Annotations}
\\-- (They/(Arthur Dempster, Nan Laird, and Donald Rubin) ; 
\\\-\hspace{70pt} pointed out that ; 
\\\-\hspace{15pt} (the method)/(The EM algorithm) had been "proposed
\\\-\hspace{15pt} many times in special circumstances" by earlier authors)
\\-- ((the method)/(The EM algorithm) ; had been proposed by;
\\\-\hspace{25pt} earlier authors ; in special circumstances) [attributed]
\\-- (earlier authors ; proposed ; (the method)/(The EM
\\\-\hspace{25pt} algorithm) ; in special circumstances) [attributed{]}
\\\smallskip\hrulefill

\textbf{Sentence FI 2 --} ``\textit{A police statement did not name the man
  in the boot, but in effect indicated the traveler was State Secretary Samuli
  Virtanen, who is also the deputy to Foreign Minister Timo Soini.}''
\smallskip\\\textbf{Annotations}
\\-- (A police/(Finnish police) statement ; did not name ;\\\-\hspace{50pt} (the man in the boot)/(Samuli Virtanen))
\\-- ((the man in the boot)/(Samuli Virtanen) ; was ; \\\-\hspace{50pt}Samuli Virtanen) [attributed]
\\-- ((the traveler)/(Samuli Virtanen) ; was ; Samuli Virtanen)\\\-\hspace{170pt} [attributed]
\\-- (Samuli Virtanen ; [is] ; State Secretary)
\\-- (Samuli Virtanen ; is ; the deputy to Foreign Minister \\\-\hspace{150pt}Timo Soini)
\\-- (Samuli Virtanen ; is ; [a] deputy)
\\-- (Timo Soini ; [is] ; Foreign Minister)
\\-- (Timo Soini ; [has] ; [a] deputy)
\\\smallskip\hrulefill

\textbf{Sentence CE 4 --} ``\textit{The International Monetary Fund, for example, saw 2017 global growth at 3.4 percent with advanced economies advancing 
1.8 percent.}''
\smallskip\\\textbf{Annotations}
\\-- (The International Monetary Fund ; saw ; 2017 global \\\-\hspace{130pt}growth ; at 3.4 percent)
\\-- (The International Monetary Fund ; saw ; advanced \\\-\hspace{50pt} economies ; advancing 1.8 percent ; [in] 2017)
\\-- (2017 global growth ; [was] ; 3.4 percent)
\\-- (advanced economies ; [advanced] ; 1.8 percent ; \\\-\hspace{138pt} [in] 2017) [attributed]
  \caption{Sample annotations from WiRe57, from four of the documents. Reformulated words are enclosed in [brackets] and coreference
    information is indicated with forward slashes and parentheses.}
  \label{fig-sample-annot}
\end{figure}

\subsection{Annotation guidelines\footnote{We share at \url{https://github.com/rali-udem/WiRe57} our annotation guidelines. We present its major points here.}}
\label{sec:orgf2ac4e9}

Extracted tuples should reflect all meaningful relationships found in the
source text. Typically, this means that there are multiple tuples for a given
sentence. A number of times, two arguments are connected in a sentence but the
relation that links them is implicit (e.g. \emph{Paris, France} ; \emph{the North Atlantic
Treaty Organization (NATO)} ; \emph{the Nature paper} or \emph{the Turing paper},
etc.). In this case, we annotate a somewhat arbitrary relationship (such as \emph{is
in}, \emph{stands for}, \emph{published in} and \emph{published by} respectively), the tokens
of which are thus inferred. This is the case for 39\% of our tuples.

Some OIE systems similarly attempt to hallucinate some or part of relations.
Notably, ClausIE wrongly extracts (\emph{New Delhi ; is ; India}), and MinIE gets
right (\emph{Paris ; is in ; France}). Ollie adds some ``be'' auxiliaries to otherwise
nominal relations, as in \emph{Barack Obama, former president of the United States,
[\ldots{}]}, which OpenIE 4 also infers. Yet, we acknowledge that most work in Open
IE rely on explicit predicate tokens as in
\cite{mesquita-schmidek-barbosa:2013:EMNLP}, and don't try to elicit relations
further. At scoring time, systems are not penalized for not finding inferred
words, or not finding inferred relations. If the whole predicate of a tuple is inferred, a
predicted tuple is scored on its token overlap with the arguments only.

We suggest ``platinum'' annotations, including inferred words, to be a very high
standard for extractors, while the gold standard for the task, recall-wise, is
based only on words found in the original sentences.

Noun phrases can be rich in elements of information. To solve the problem
of finding the granularity level to use when including argument NPs, we
extract two tuples, one as generic as possible and the other as specific as
possible, for the same relation. Adjectives and other elements of meaning
that can be easily separated from the noun phrase to create other tuples
are so split. Only elements that cannot be separated become part of the most specific
noun phrase.

For instance, the sentence \emph{``Solo Piano is a great album of classical piano
compositions}'' would yield 3 tuples : the split adjective (\emph{Solo Piano ; is
; great}), the generic (\emph{Solo Piano ; is ; [an] album}) and the specific
(\emph{Solo Piano ; is ; [an] album of classical piano compositions}).

When predicates contain nouns or other elements (e.g. \emph{Tokyo is the
capital of Japan.}), we annotate the richer relationship (\emph{Tokyo ; is the
capital of ; Japan}) rather than the more basic (\emph{Tokyo ; is ; the capital
of Japan}). This allows tuple relations to be more meaningful, and more
easily compared, clustered, and aggregated with other relations. This also
is in line with ReVerb.

Like ClausIE and other extractors since, we split conjunctions : \emph{``Andrea
lived in both Poland and Italy}'' yields both (\emph{Andrea ; lived in ;
Poland}) and (\emph{Andrea ; lived in ; Italy}).

\subsection{Resource}
\label{sec:org2c01bf4}

A sample of annotations is pictured in Figure \ref{fig-sample-annot}. The
occurring frequency of various phenomena is presented in Table \ref{tab:org109a930}.
Our resource is comprised of 343 relational facts (or tuples), three quarters of them
binary relations. One in five have three arguments, sometimes ``two objects'' as
in (\emph{This performance ; has made ; some economists ; optimistic}) or more
frequently a complement as in (\emph{His parents ; had to flee ; from Hungary ;
during World War II}). Five percent of them have four arguments or more : for
instance (\emph{Tokyo ; ranked ; third ; in the International Financial Centres
Development IndexEdit~; twice}) and (\emph{The International Monetary Fund ; saw ;
advanced economies ; advancing 1.8 percent ; [in] 2017}).

We found (and resolved) anaphoric phrases in more than half the
tuples, as in (\emph{Emperor Meiji~; moved} ; \emph{his/(Emperor Meiji's) seat} ; \emph{to
(the city)/Tokyo} ; \emph{from the old capital of Kyoto ; in 1868}). The released
dataset contains the raw and anaphora-resolved argument spans.

When solely extracting words from the sentence would not yield clear factual
tuples, we reworded or adapted the text into more explicit statements. In this
case, we explicitly marked the changed (or added) words as inferred (they are bracketed in Figure
\ref{fig-sample-annot}). For instance in sentence CE 4, the relation ``[advanced]'' was reformulated from the sentence word
``\textit{advancing}'', and the word [in] was added before ``2017''. In the resource, each token is
accompanied by its index in the sentence if it comes from it, or the ``inferred''
mark. Inferred words represent 13\% of the lot but affect 54\% of the tuples.

\subsection{Inter-annotator agreement}

\begin{table}[hbt]
\centering
\begin{center}
\begin{tabular}{lrrrr}
 & \# tokens & 1$\leftrightarrow$2 & 1$\leftrightarrow$R & 2$\leftrightarrow$R\\
\hline
Sentence 1 & 24 & 84.4 & 90.6 & 93.8\\
Sentence 2 & 19 & 98.7 & 98.7 & 100\\
Sentence 3 & 33 & 78.0 & 90.9 & 85.6\\
\hline
\textbf{Average} &  & \textbf{85.2} & \textbf{92.8} & \textbf{91.9}\\
\end{tabular}
\end{center}
\caption{\label{tab:IAA}\textbf{Inter-annotator agreement.} Percentage of agreement on the labelling of each sentence token as belonging to 4 classes. Each annotator's original production differs only slightly from the agreed-on result (columns 1$\leftrightarrow$R and 2$\leftrightarrow$R), and the disagreement between both annotators is slightly larger (column 1$\leftrightarrow$2). The average is computed token-wise.}
\end{table}

As mentioned in section~\ref{sec:org85568e7}, a qualitatively high agreement was reached after the merging of preliminary annotations and deliberation over the guidelines' items. After the guidelines were fully settled, three additional sentences from one of the documents were annotated by two annotators (1 and 2) in order to \textit{quantitatively} measure inter-annotator agreement. 
Afterward, annotation discrepancies were resolved in cases of disagreement to produce a merged reference (R). 
Here, we report the agreement between the two original annotations (1$\leftrightarrow$2), and between each original annotation and the merged reference (1$\leftrightarrow$R and 2$\leftrightarrow$R). 

Comparing triples can become quite tricky for many reasons, including missing complements, overlapping spans, etc. We therefore resorted to another scheme, where we reframe the annotation task as taking each annotated token and classifying it as either belonging or not belonging to each of 4~classes (subject, relation, object, or complementary argument). These classifications can be trivially derived from the triples produced beforehand. For instance, a triple $(t_1~t_2;t_3;t_4~t_5)$ implies that the annotator classified tokens $t_1$ and $t_2$ as belonging to the subject class. It then becomes possible to measure an agreement percentage on the
full binary labelling grid (obtained automatically from the long-form annotations). We believe the resulting figures (shown in Table~\ref{tab:IAA}) aptly reflect the level of overall agreement between the annotators, despite the minimal sample size. We measure an overall inter-annotator agreement (1$\leftrightarrow$2) of 85.2\% for the three sentences.

\begin{table*}[htb!]
\setlength{\tabcolsep}{1.2mm}
\centering
\makebox[\textwidth][c]{
\begin{tabular}{l|ccc|cc|ccc}
 & \small{Extractions} & \small{Matches} & \thead{Exact \\ matches} & \thead{Prec. of \\ matches} & \thead{Recall of \\ matches} & Prec. & Recall & F1\\
\hline
ReVerb \cite{Fader:2011:IRO:2145432.2145596}
& 79  & 54  & 13 & .83 & .77 & \textbf{.569} & .121 & .200\\
Ollie \cite{ollie-emnlp12}
& 145 & 74  & 8  & .73 & .81 & .347 & .175 & .239\\
ClausIE \small{\cite{DelCorro:2013:CCO:2488388.2488420}} 
& 223 & 121 & \textbf{24} & .74 & .84 & .401 & .298 & .342\\
Stanford \cite{2015angeli-openie}
& 371 & 99  & 2  & .79 & .65 & .210 & .188 & .198\\
OpenIE 4 \cite{Mausam:2016:OIE:3061053.3061220} 
& 101 & 74  & 5  & .68 & .84 & .501 & .182 & .267\\
PropS  \cite{DBLP:journals/corr/StanovskyFDG16} 
& 184 & 69  & 0  & .59 & .80 & .222 & .162 & .187\\
MinIE \cite{D17-1278} 
& 252 & \textbf{134} & 10 & .75 & .83 & .400 & \textbf{.323} & \textbf{.358}\\
\end{tabular}} 
\caption{\label{tab:org9aab7f6}
Performance of available OpenIE systems (in chronological order) on our reference. Precision and recall are computed at the token level. Systems with lower precision of matches are penalized for producing overlong tuples. High precision and recall of matches overall show that our matching function (one shared word in each of the first three parts) works correctly. Inferred words are required for exact matches.}
\end{table*}

Qualitatively, one annotator steered close to the sentence syntax, sometimes
missing some of the meaning obscured by long-winded formulations. The other
annotator tended to be overly specific, including some
non-essential complements, and making longer-ranged inferences that fall out of
the scope of this task. Some possessive and passive constructions were also
overlooked.

\section{Evaluation of Existing Systems}
\label{sec:org231c554}

\subsection{Scorer}
\label{sec:org4b13ea5}

An important step when measuring extractors' performances is the scoring
process. Matching a system's output to a reference is not trivial.  As detailed
in Section \ref{sec:org4aff63d}, because it didn't penalize overlong
extractions, we could game the basic evaluation method of the QA-SRL OIE
benchmark with a trivial extractor.

Our scorer computes precision and recall of a system's predicted tuples at the
token level. Precision is, briefly put, the proportion of extracted words that
are found in the reference. Recall is the proportion of reference words found
in the systems' predictions.

More formally, let \(G = \{g_1, g_2, \dots, g_N\}\) be the gold tuples, and
\(T_{\text{sys}} = \{t_1, t_2, \dots, t_n\}\) a system's extractions. We denote
the parts of a tuple \(t = (t^{a_1} ; t^r ; t^{a_2} ; t^{a_3} ; \dots) =
(t^{p_k})_{k\in[1,6]}\), where \(p_1\) is the first argument, \(p_2\) is the
relation, etc., up to \(p_6\) the fifth argument when it exists (no reference
tuple contains more than 5 arguments). Let \(t_i^{p} \cap g_j^{p}\) be the subset
of words shared by parts \(t_i^{p}\) and \(g_j^{p}\), where parts are considered as
bags of words. The length of a tuple is the sum of lengths of its parts,
i.e. \(|t_i| = |t_i^{a_1}| + |t_i^r| + |t_i^{a_2}| + |t_i^{a_3}|+\dots =
\sum_k |t_i^{p_k}|\).

A predicted tuple \(t_i\) may match a reference tuple \(g_j\) from the same
sentence if they share at least one word from each of the relation, first and
second arguments, that is iff \((w_{a_1}, w_r, w_{a_2})\) exist such
that \(w_{1} \in g_{j}^{a_1} \cap t_{i}^{a_1}, w_r \in g_{j}^{r} \cap t_{i}^{r}\)
and \(w_{2} \in g_{j}^{a_2} \cap t_{i}^{a_2}\).

For all tuple pairs that may match, we have the matching scores: 
\begin{align*}
    \operatorname{precision}(t_i, g_j) & = \frac{\sum_k|t_i^{p_k} \cap g_j^{p_k}|}{|t_i|} \\
    \operatorname{recall}(t_i, g_j) & = \frac{\sum_k|t_i^{p_k} \cap g_j^{p_k}|}{|g_j|}\\
    F_1 & = \frac{2\ p\ r}{p+r}.
\end{align*}

We match predicted tuples with reference ones by greedily removing from the potential match pool the pair with maximum \(F_1\) score, until no remaining tuples match. Let \(m(.)\) be the matching function such that \(t_i\) matches with \(g_{m(i)}\) (and conversely \(t_{m(j)}\) matches \(g_j\)), assuming that \(|t_i \cap g_{m(i)}| = 0\) if there is no match for \(t_i\).

Hence, the overall performance metrics of an extractor are its token-weighted precision and recall over all tuples, i.e. 
\begin{align*}
\operatorname{precision}_{\text{sys}} & = \frac{{\displaystyle\sum_{i}^n}\left(\sum_k|t_i^{p_k} \cap g_{m(i)}^{p_k}|\right)}{\sum_{i}^n|t_i|}\\
\operatorname{recall}_{\text{sys}} & = \frac{{\displaystyle\sum_{j}^N}\left(\sum_k|t_{m(j)}^{p_k} \cap g_{j}^{p_k}|\right)}{\sum_{j}^N|g_j|}\\
{F_1}_{\text{sys}} & = \frac{2\ p_{\text{sys}}\ r_{\text{sys}}}{p_{\text{sys}}+r_{\text{sys}}}.
\end{align*}

To avoid penalizing systems for not finding them, neither the words annotated
as inferred, nor the coreference information are used in this evaluation (\(g_j\)
is the non-resolved version of the tuple, and inferred words are not included
in recall denominators). Future work can look into evaluating OIE systems that
mean to resolve anaphoras.

\subsection{Results}
\label{sec:orgf4eff85}

In order to experiment with the 7 systems used in this paper, we
bundled them as a web service. A client application
need only submit a sentence and a list of OIE system names to perform
extraction. All tuples are in turn served as uniform JSON objects, no matter the OIE system used. This
facilitates the development of clients, shielded from the
various tuple formats, coding languages,
and other quirks of the OIE systems. It also allowed us to
visualize the tuples using a web application (see
Figure~\ref{fig-buddleia}). Moreover, because the various extractors run as servers,
they load their respective resources only once, when the service is launched, 
and are then always quick to respond to a given extraction task (a few seconds). Otherwise, 
the user would have had to wait a few minutes for the resources to load each time when querying the extractors.

While creating such a framework is a significant
effort, it ultimately saved us a lot of time when writing the clients. It
also provided a common frame of reference for all collaborators in our
lab. Typically, we used the default configuration for each OIE system, but
we tweaked the available flags in order to favor exhaustiveness, when such
flags were present and properly documented. When additional information did
not fit into a traditional tuple (arg1; rel; arg2), e.g. MinIE's
quantities, we resorted to simple schemes to faithfully cast that
information into a tuple.

Table \ref{tab:org9aab7f6} details the performance of available OIE systems against our
reference. MinIE produces a large number of correct tuples, and performs best, especially recall-wise. The conservative
choices made by ReVerb achieve a relatively high precision, though it lacks
in comprehensiveness. Ollie improves recall over ReVerb, and Open IE 4 improves precision over Ollie. Stanford Open IE produces a very large number of tuples, hindering its precision (it is possible that limiting its verboseness through configuration would improve this). 
\section{Conclusion}
\label{sec:orgad750a7}

In this paper, we set out to create additional resources useful to researchers in Open Information Extraction. We distribute these resources freely.

Primarily, we provide a manually crafted, tentative reference for the task. It consists of 343 manually extracted facts, including some implicit relations, over 57 sentences. A quarter of them are $n$-ary relations and coreference information is included in over half of them. We believe that such a benchmark is valuable because it offers a common frame of reference allowing OIE systems to be tested and compared fairly, a task we carried out on 7~OIE systems. This also entailed the creation of a scoring algorithm and program, which we release along with the data. We assess the ReVerb, Ollie, ClausIE, Stanford Open IE, OpenIE 4, PropS, and MinIE
systems against our reference, using a fine-grained token-level scorer. We find the MinIE system to perform best.

Naturally, such an annotation effort requires one to attempt to ``pin down'' the task of OIE by confronting real-life data. We provide guidelines that propose such a definition. While by no means definitive or exhaustive, these guidelines have at least the merit of being sufficiently clear to yield an annotated dataset with a reasonable inter-annotator agreement. At the same time, we believe they are not too overwrought, and rather invite further contributions by other researchers. The thorniest issues are the fine line between useful reformulation of information to a canonical form and ill-advised inference, and how to trim and annotate complex noun-phrase arguments. These difficulties can affect the manual annotation process, and, interestingly, are also likely to arise when building OIE systems, which is the ultimate goal in this research field after all.

\bibliographystyle{acl_natbib_nourl}
\bibliography{bibliography.bib}{}
\end{document}